# Updating Probabilities

Peter D. Grünwald
CWI, P.O. Box 94079
1090 GB Amsterdam
pdg@cwi.nl
www.cwi.nl/~pdg

Joseph Y. Halpern
Cornell University
Ithaca, NY 14853
halpern@cs.cornell.edu
www.cs.cornell.edu/home/halpern

## Abstract

As examples such as the Monty Hall puzzle show, applying conditioning to update a probability distribution on a "naive space", which does not take into account the protocol used, can often lead to counterintuitive results. Here we examine why. A criterion known as CAR ("coarsening at random") in the statistical literature characterizes when "naive" conditioning in a naive space works. We show that the CAR condition holds rather infrequently. We then consider more generalized notions of update such as Jeffrey conditioning and minimizing relative entropy (MRE). We give a generalization of the CAR condition that characterizes when Jeffrey conditioning leads to appropriate answers, but show that there are no such conditions for MRE. This generalizes and interconnects previous results obtained in the literature on CAR and MRE.

## 1   INTRODUCTION

Suppose an agent represents her uncertainty about a domain using a probability distribution. At some point, she receives some new information about the domain. How should she update her distribution in the light of this information? *Conditioning* is by far the most common method in case the information comes in the form of an event. However, there are numerous well-known examples showing that naive conditioning can lead to problems. We give just two of them here.

**Example 1.1:** The *Monty Hall puzzle* [Mosteller 1965; vos Savant 1990]: Suppose that you're on a game show and given a choice of three doors. Behind one is a car; behind the others are goats. You pick door 1. Before opening door 1, Monty Hall, the host (who knows what is behind each door) opens door 3, which has a goat. He then asks you if you still want to take what's behind door 1, or to take what's behind door 2 instead. Should you switch? Assuming that, initially, the car was equally likely to be behind each of the doors, naive conditioning suggests that, given that it is not behind door 3, it is equally likely to be behind door 1 and door 2. Thus, there is no reason to switch. However, another argument suggests you should switch: if a goat is behind door 1 (which happens with probability 2/3), switching helps; if a car is behind door 1 (which happens with probability 1/3), switching hurts. Which argument is right? ∎

**Example 1.2:** The *three-prisoners puzzle* [Bar-Hillel and Falk 1982; Gardner 1961; Mosteller 1965]: Of three prisoners $a$, $b$, and $c$, two are to be executed, but $a$ does not know which. Thus, $a$ thinks that the probability that $i$ will be executed is 2/3 for $i \in \{a, b, c\}$. He says to the jailer, "Since either $b$ or $c$ is certainly going to be executed, you will give me no information about my own chances if you give me the name of one man, either $b$ or $c$, who is going to be executed." But then, no matter what the jailer says, naive conditioning leads $a$ to believe that his chance of execution went down from 2/3 to 1/2. ∎

Why does naive conditioning give the wrong answer in these examples? As argued in [Halpern and Tuttle 1993; Shafer 1985], the real problem is that we are not conditioning in the right space. If we work in a larger "sophisticated" space, where we take the protocol used by Monty (in Example 1.1) and the jailer (in Example 1.2) into account, conditioning does deliver the right answer. Roughly speaking, the sophisticated space consists of all the possible sequences of events that could happen (for example, what Monty would say in each circumstance, or what the jailer would say in each circumstance), with their probability.[1] However, working in the sophisticated space has problems too. For one thing, it is not always clear what the relevant probabilities in the sophisticated space are. For example, what is the probability that the jailer says $b$ if $a$ is to be executed? Indeed, in some cases, it is not even clear what the elements of the larger space are. Moreover, even when the elements and the relevant probabilities are known, the size of the sophisticated space may become an issue, as the following example shows.

---

[1] The notions of "naive space" and "sophisticated space" will be formalized in Section 2. This introduction is meant only to give an intuitive feel for the issues.

**Example 1.3:** Suppose that a world describes which of 100 people have a certain disease. A world can be characterized by a tuple of 100 0s and 1s, where the $i$th component is 1 iff individual $i$ has the disease. There are $2^{100}$ possible worlds. Further suppose that the "agent" in question is a computer system. Initially, the agent has no information, and considers all $2^{100}$ worlds equally likely. The agent then receives information that is assumed to be true about which world is the actual world. This information comes in the form of statements like "individual $i$ is sick or individual $j$ is healthy" or "at least 7 people have the disease". Each such statement can be identified with a set of possible worlds. For example, the statement "at least 7 people have the disease" can be identified with the set of tuples with at least 7 1s. For simplicity, assume that the agent is given information saying "the actual world is in set $U$", for various sets $U$. Suppose at some point the agent has been told that the actual world is in $U_1, \ldots, U_n$. Then, after doing conditioning, the agent has a uniform probability on $U_1 \cap \ldots \cap U_n$.

But how does the agent keep track of the worlds it considers possible? It certainly will not explicitly list them; there are simply too many. One possibility is that it keeps track of what it has been told; the possible worlds are then the ones consistent with what it has been told. But this leads to two obvious problems: checking for consistency with what it has been told may be hard, and if it has been told $n$ things for large $n$, remembering them all may be infeasible. In situations where these two problems arise, an agent may not be able to condition appropriately. ∎

Example 1.3 provides some motivation for working in the smaller, more naive space. Examples 1.1 and 1.2 show that this is not always appropriate. Thus, an obvious question is when it is appropriate. It turns out that this question is highly relevant in the statistical areas of *selectively reported data* and *missing data*. Originally studied within these contexts [Rubin 1976; Dawid and Dickey 1977], it was later found that it also plays a fundamental role in the statistical work on *survival analysis* [Kleinbaum 1999]. Building on previous approaches, Heitjan and Rubin [1991] presented a necessary and sufficient condition for when conditioning in the "naive space" is appropriate. Nowadays this so-called *CAR (Coarsening at Random)* condition is an established tool in survival analysis. (See [Gill, van der Laan, and Robins 1997; Nielsen 1998] for overviews.) We examine this criterion in our own, rather different context, and show that it applies rather rarely.

We then show that the situation is even worse if the information does not come in the form of an event. For that case, several generalizations of conditioning have been proposed. Perhaps the best known are *Jeffrey conditioning* [Jeffrey 1968] (also known as *Jeffrey's rule*) and *Minimum Relative Entropy (MRE) Updating* [Kullback 1959; Shore and Johnson 1980] (also known as *cross-entropy*). Jeffrey conditioning is a generalization of ordinary conditioning; MRE updating is a generalization of Jeffrey conditioning.

We show that Jeffrey conditioning, when applicable, can be justified under an appropriate generalization of the CAR condition. Although it has been argued, using mostly axiomatic characterizations, that MRE updating (and hence also Jeffrey conditioning) is, when applicable, the *only* reasonable way to update probability (see, e.g., [Csiszár 1991; Shore and Johnson 1980]), it is well known that there are situations where applying MRE leads to paradoxical, highly counterintuitive results [Seidenfeld 1986; van Fraassen 1981].

**Example 1.4:** Consider the *Judy Benjamin* problem [van Fraassen 1981]: Judy is lost in a region that is divided into two halves, Blue and Red territory, each of which is further divided into Headquarters Company area and Second Company area. A priori, Judy considers it equally likely that she is in any of these four quadrants. She contacts her own headquarters by radio, and is told "I can't be sure where you are. If you are in Red territory, the odds are 3:1 that you are in HQ Company area ..." At this point the radio gives out. MRE updating on this information leads to a distribution where the posterior probability of being in Blue territory is greater than $1/2$. Indeed, if HQ had said "If you are in Red territory, the odds are $\alpha$ : 1 that you are in HQ company area ...", then for all $\alpha \neq 1$, according to MRE updating, the posterior probability of being in Blue territory is always greater than $1/2$. ∎

In [Grove and Halpern 1997], a "sophisticated space" is provided where conditioning gives what is arguably the more intuitive answer in the Judy Benjamin problem, namely that if HQ sends a message of the form "if you are in Red territory, then the odds are $\alpha$ : 1 that you are in HQ company area" then Judy's posterior probability of being in each of the two quadrants in Blue remains at $1/4$. Seidenfeld [1986], strengthening results of Friedman and Shimony [1971], showed that there is *no* sophisticated space in which conditioning will give the same answer as MRE in this case. (See also [Dawid 2001] for similar results along these lines.) We strengthen these results by showing that, even in a class of much simpler situations (where Jeffrey conditioning cannot be applied), using MRE in the naive space corresponds to conditioning in the sophisticated space in essentially only trivial cases. These results taken together show that *generally speaking, working with the naive space, while an attractive approach, is likely to give highly misleading answers.* That is the main message of this paper.

We remark that, although there are certain similarities, our results are quite different in spirit from the well-known results of Diaconis and Zabell [1986]. They considered when a posterior probability could be viewed as the result of conditioning a prior probability on some larger space. By way of contrast, we have a fixed larger space in mind (the "sophisticated space"), and are interested in when conditioning in the naive space and the sophisticated space agree.

It is also worth stressing that the distinction between the naive and the sophisticated space is entirely unrelated to the philosophical view that one has of probability and how one should do probabilistic inference. For example, the probabilities in the Monty Hall puzzle can be viewed as the participant's subjective probabilities about the location of the car and about what Monty will say under what circumstances; alternatively, they can be viewed as "frequentist" probabilities, inferred from watching the Monty Hall show on television for many weeks and then setting the probabilities equal to observed frequencies. The problem we address occurs both from a frequentist and from a subjective stance.

The rest of this paper is organized as follows. In Section 2 we formalize the notion of naive and sophisticated spaces. In Section 3, we consider the case where the information comes in the form of an event. We describe the CAR condition and show by example how rarely it applies. In Section 4 we consider the case where the information is not in the form of an event. We first consider situations where Jeffrey conditioning can be applied. We show that Jeffrey conditioning in the naive space gives the appropriate answer iff a generalized CAR condition holds. We then show that, except in trivial cases, applying MRE in the naive space does not give the appropriate answer. We conclude with some discussion of the implication of these results in Section 5.

## 2   NAIVE VS. SOPHISTICATED SPACES

Our formal model is a special case of the multi-agent systems framework [Halpern and Fagin 1989], which is essentially the same as that used in [Friedman and Halpern 1997] to model belief revision. We assume that there is some external world in a set $W$, and an agent who makes observations or gets information about that world. We can describe the situation by a pair $(w, l)$, where $w \in W$ is the actual world, and $l$ is the agent's *local state*, which essentially characterizes her information. $W$ is what we called the "naive space" in the introduction. For the purposes of this paper, we assume that $l$ has the form $\langle o_1, \ldots, o_n \rangle$, where $o_j$ is the observation that the agent makes at time $j$, $j = 1, \ldots, n$. This representation implicitly assumes that the agent remembers everything she has observed (since her local state encodes all the previous observations). Thus, we ignore memory issues here. We also ignore computa-tional issues, just so as to be able to focus on when conditioning is appropriate.

A pair $(w, \langle o_1, \ldots, o_n \rangle)$ is called a global state. A *run* is a function from time to global states. Thus, if $r$ is a run, then $r(0), r(1), \ldots$ is a sequence of global states that, roughly speaking, is a complete description of what happens over time in one possible execution of the system. If $r(m) = (w, \langle o_1, \ldots, o_m \rangle)$, then we let $r_W(m) = w$ and $r_O(m) = \langle o_1, \ldots, o_m \rangle$. For simplicity, in this paper, we assume that the state of the world does not change over time, so that $r_W$ is a constant function. The "sophisticated space" is the set of all possible runs.

In the Monty Hall puzzle, the naive space has three worlds, representing the three possible locations of the car. The sophisticated space describes what Monty would have said in all circumstances (i.e., Monty's *protocol*) as well as where the car is. The three-prisoners puzzle is treated in detail in Example 2.1 below. While in these cases the sophisticated space is still relatively simple, this is no longer the case for the Judy Benjamin puzzle. Although the naive space has only four elements, constructing the sophisticated space involves considering all the things that HQ could have said, which is far from clear, and the conditions under which HQ says any particular thing.

In general, not only is it not clear what the sophisticated space is, but the need for a sophisticated space and the form it must take may become clear only after the fact. For example, in the Judy Benjamin problem, before contacting headquarters, Judy would almost certainly not have had a sophisticated space in mind (even assuming she was an expert in probability), and could not have known the form it would have to take until after hearing headquarter's response.

In any case, if the agent has a prior probability on the set $\mathcal{R}$ of possible runs in the sophisticated space, after hearing or observing $\langle o_1, \ldots, o_k \rangle$, she can condition, to get a posterior on $\mathcal{R}$. Formally, the agent is conditioning her prior on the set $\mathcal{R}[\langle o_1, \ldots, o_k \rangle]$ of runs where her local state at time $k$ is $\langle o_1, \ldots, o_k \rangle$.

Clearly the agent's probability $\Pr$ on $\mathcal{R}$ induces a probability $\Pr_W$ on $W$ by marginalization. We are interested in whether the agent can compute her posterior on $W$ after observing $\langle o_1, \ldots, o_k \rangle$ in a relatively simple way, without having to work in the sophisticated space.

**Example 2.1:** Consider the three-prisoners puzzle in more detail. Here the naive space is $W = \{w_a, w_b, w_c\}$, where $w_x$ is the world where $x$ is not executed. We are only interested in runs of length 1, so $n = 1$. The set $O$ of observations (what agent can be told) is $\{\{w_a, w_b\}, \{w_a, w_c\}\}$. Here "$\{w_a, w_b\}$" corresponds to the observation that either $w_a$ or $w_b$ will *not* be executed (i.e., the jailer saying "$c$ will be executed"); similarly, $\{w_a, w_c\}$ corresponds to the jailer saying "b will be executed". The sophisticated space consists of the four runs of the form ($r(0) = (w_x, \langle \rangle); r(1) = (w_x, (\{w_x, w_y\}))$) where $x \neq y$ and $\{w_x, w_y\} \neq \{w_b, w_c\}$ (since the jailer will not tell $a$ that he will not be executed). According to the story, the prior $\text{Pr}_W$ in the naive space has $\text{Pr}_W(w) = 1/3$ for $w \in W$. The full prior Pr on $\mathcal{R}$ is not completely specified by the story, and will be discussed further in Example 3.3. ∎

## 3   THE CAR CONDITION

A particularly simple setting is where the agent observes or learns that the external world is in some set $U \subseteq W$. For simplicity, we assume that the agent makes only one observation, and makes it at the first step of the run.[2] Thus, the set $O$ of possible observations consists of subsets of $W$. However, $O$ is not necessarily $2^W$. Some subsets may never be observed. For example, in Example 2.1, $a$ is never told that he will be executed, so $\{w_b, w_c\}$ is not observed. We assume that the agent's observations are accurate, in that if the agent observes $U$ in a run $r$, then the actual world in $r$ (i.e., $r_W(0)$) is in $U$. In Example 2.1, accuracy is enforced by the requirement that $r(1)$ has the form ($w_x, (\{w_x, w_y\})$).

The observation or information obtained does not have to be exactly of the form "the actual world is in $U$". It suffices that it is equivalent to such a statement. This is the case in both the Monty Hall puzzle and the three-prisoners puzzle. For example, in the three-prisoners puzzle, being told that $b$ will be executed is essentially equivalent to observing $\{w_a, w_c\}$ (either $a$ or $c$ will not be executed).

In this setting, we can ask whether, after observing $U$, the agent can compute her posterior on $W$ by conditioning on $U$. Roughly speaking, this amounts to asking whether observing $U$ is the same as discovering that $U$ is true. This may not be the case in general—observing or being told $U$ may carry more information than just the fact that $U$ is true. For example, if for some reason $a$ knows that the jailer would never say $c$ if he could help it (so that, in particular, if $b$ and $c$ will be executed, then he will definitely say $b$), then hearing $c$ (i.e., observing $\{w_a, w_b\}$) tells $a$ much more than the fact that the true world is one of $w_a$ or $w_b$. It says that the true world must be $w_b$ (for if the true world were $w_a$, the jailer would have said $b$).

For ease of exposition, in the remainder of this paper we assume that $W$ and $\mathcal{R}$ are finite, and that all nonempty subsets of $\mathcal{R}$ are measurable. Moreover, whenever we speak of a distribution Pr over $\mathcal{R}$, we implicitly assume that the probability of any set on which we condition is strictly greater than 0. Let $\mathcal{R}[U]$ consist of all runs in $\mathcal{R}$ where the true world is in $U$ (i.e., $r_W(0) \in U$). As before, let $\mathcal{R}[\langle U \rangle]$ consist of all runs where the agent observes $U$ at the first step. Let Pr be a prior on $\mathcal{R}$ and let $\text{Pr}' = \text{Pr}(\cdot \mid \mathcal{R}[\langle U \rangle])$ be the posterior after observing $U$. Thus, we are interested in knowing whether $\text{Pr}'_W(V) = \text{Pr}_W(V \mid U)$; that is, whether the posterior on $W$ induced by $\text{Pr}'$ can be computed from the prior on $W$ by conditioning on the observation. (Example 3.3 below gives a concrete case.) We stress that Pr and $\text{Pr}'$ are distributions on $\mathcal{R}$, while $\text{Pr}_W$ and $\text{Pr}'_W$ are distributions on $W$ (obtained by marginalization from Pr and $\text{Pr}'$, respectively).

The following simple proposition says that this can be done iff conditioning on $U$ is equivalent to conditioning on observing $U$.

**Proposition 3.1:** *Let* $\text{Pr}' = \text{Pr}(\cdot \mid \mathcal{R}[\langle U \rangle])$. *Then* $\text{Pr}'_W = \text{Pr}_W(\cdot \mid U)$ *iff* $\text{Pr}(\mathcal{R}[V] \mid \mathcal{R}[U]) = \text{Pr}(\mathcal{R}[V] \mid \mathcal{R}[\langle U \rangle])$ *for all* $V \subseteq W$.

Now the obvious question is when $\text{Pr}(\mathcal{R}[V] \mid \mathcal{R}[U]) = \text{Pr}(\mathcal{R}[V] \mid \mathcal{R}[\langle U \rangle])$. The CAR condition characterizes this. It is best stated in terms of random variables. Let $X_W$ and $X_O$ be two random variables on $\mathcal{R}$, where $X_W$ is the actual world and $X_O$ is the first event observed. Thus, $X_W(r) = r_W(0)$ and $X_O(r) = U$ if $r_O(1) = \langle U \rangle$. Note that $\mathcal{R}[U]$ is $X_W \in U$ (that is, $\mathcal{R}[U] = \{r : X_W(r) \in U\}$) and $\mathcal{R}[\langle U \rangle]$ is $X_O = U$.

**Theorem 3.2** *[Gill, van der Laan, and Robins 1997] Fix a probability* Pr *on* $\mathcal{R}$ *and a set* $U \subseteq W$ *The following are equivalent:*

(a) *If* $\text{Pr}(X_O = U) > 0$, *then* $\text{Pr}(X_W = w \mid X_O = U) = \text{Pr}(X_W = w \mid X_W \in U)$ *for all* $w \in U$.

(b) *The event* $X_W = w$ *is independent of the event* $X_O = U$ *given* $X_W \in U$, *for all* $w \in U$;

(c) $\text{Pr}(X_O = U \mid X_W = w) = \text{Pr}(X_O = U \mid X_W \in U)$ *for all* $w \in U$ *such that* $\text{Pr}(X_W = w) > 0$;

(d) $\text{Pr}(X_O = U \mid X_W = w) = \text{Pr}(X_O = U \mid X_W = w')$ *for all* $w, w' \in U$ *such that* $\text{Pr}(X_W = w) > 0$ *and* $\text{Pr}(X_W = w') > 0$.

The proof of this and all other results can be found in the full paper (available at www.cwi.nl/~pdg).

The first condition in Theorem 3.2 just says that $\text{Pr}(\mathcal{R}[\{w\}] \mid \mathcal{R}[\langle U \rangle]) = \text{Pr}(\mathcal{R}[\{w\}] \mid \mathcal{R}[U])$ for all $w \in W$. Given that $W$ is finite, this is clearly equivalent to the desired condition $\text{Pr}(\mathcal{R}[V] \mid \mathcal{R}[\langle U \rangle]) = \text{Pr}(\mathcal{R}[V] \mid \mathcal{R}[U])$. The third and fourth conditions justify the name "coarsening at random". Intuitively, first some world $w \in W$ is realized, and then some "coarsening mechanism" decides which event $U \subseteq W$ such that $w \in U$ is revealed to the agent. The event $U$ is called a "coarsening" of $w$. The third and fourth conditions effectively say that the probability that $w$ is coarsened to $U$ is the same for all $w \in U$. This means that the "coarsening mechanism" is such that the probability of observing $U$ is not affected by the specific value of $w \in U$ that was realized.

---

[2] We can easily extend the results to allow for multiple observations at many steps.

The CAR condition explains why conditioning in the naive space is not appropriate in the Monty Hall puzzle or the three-prisoners puzzle. We consider the three-prisoners puzzle in detail; a similar analysis applies to Monty Hall.

**Example 3.3:** In the three-prisoners puzzle, what is $a$'s prior distribution $\Pr$ on $\mathcal{R}$? In Example 2.1 we assumed that the marginal distribution $\Pr_W$ over $W$ is uniform. Apart from this, $\Pr$ is unspecified. Now suppose that $a$ observes $\{w_a, w_c\}$ ("the jailer says $b$"). Naive conditioning would lead $a$ to adopt the distribution $\Pr_W(\cdot \mid \{w_a, w_c\})$. This distribution satisfies $\Pr_W(w_a \mid \{w_a, w_c\}) = 1/2$. Sophisticated conditioning leads $a$ to adopt the distribution $\Pr' = \Pr(\cdot \mid \mathcal{R}[\langle \{w_a, w_c\} \rangle])$. By part (d) of Theorem 3.2, naive conditioning is appropriate (i.e., $\Pr'_W = \Pr_W(\cdot \mid \{w_a, w_c\})$ only if the jailer is equally likely to say $b$ in both worlds $w_a$ and $w_c$. Since the jailer must say that $b$ will be executed in world $w_c$, it follows that $\Pr(X_O = \{w_a, w_c\} \mid X_W = w_c) = 1$. Thus, conditioning is appropriate only if the jailer's protocol is such that he definitely says $b$ in $w_a$, i.e., even if both $b$ and $c$ are executed. But if this is the case, when the jailer says $c$, conditioning $\Pr_W$ on $\{w_a, w_b\}$ is *not* appropriate, since then $a$ knows that he will be executed. The world cannot be $w_a$, for then the jailer would have said $b$. ∎

So when does the CAR condition hold? There is only one simple situation where it is guaranteed to hold. Roughly speaking, this is when the observations are pairwise disjoint. Given a system $\mathcal{R}$, let $O = \{U_1, \ldots, U_n\}$ be the set of observations made in $\mathcal{R}$. Let $V_i$ be the set of worlds where $U_i$ is observed; that is $V_i = \{X_W(r) : X_O(r) = U_i, r \in \mathcal{R}\}$, for $i = 1, \ldots, n$. Since we have assumed that observations are accurate, we must have that $V_i \subseteq U_i$. Let $O^{\mathcal{R}} = \{V_1, \ldots, V_n\}$. If the sets in $O^{\mathcal{R}}$ are pairwise disjoint, then for each probability distribution $\Pr$ on $\mathcal{R}$ and each world $w \in V_j$ such that $\Pr(X_W = w) > 0$, it must be the case that $\Pr(X_O = U_j \mid X_W = w) = 1$. Thus, part (d) of Theorem 3.2 applies. Note that, if the sets in $O$ are pairwise disjoint, then the sets in $O^{\mathcal{R}}$ must also be pairwise disjoint. Whenever the set $O^{\mathcal{R}}$ does *not* consist of pairwise disjoint subsets of $W$, one can construct distributions $\Pr$ over $\mathcal{R}$ such that the CAR condition does not hold. Summarizing:

**Proposition 3.4:** *The CAR condition holds for all distributions $\Pr$ over $\mathcal{R}$ if and only if $O^{\mathcal{R}}$ consists of pairwise disjoint subsets of $W$.*

Note that the sets in $O^{\mathcal{R}}$ are pairwise disjoint iff $X_O$ can be viewed as a function on $W$ (i.e., its value in a run $r$ is completely determined by $r_W(0)$).

Are there other cases (combinations of $W$, $O$ and distributions over $\mathcal{R}$) when CAR holds? There are, but they are somewhat special. Although we have not bothered to try to get a complete characterization of when CAR holds— this involves stating a number of linear equalities that must hold, and does not give much insight—the following examples show that, in general, it can be very difficult to satisfy CAR.

**Example 3.5:** Suppose that $O = \{U_1, U_2\}$, and both $U_1$ and $U_2$ are observed with positive probability. (This is the case for both Monty Hall and the three-prisoners puzzle.) Then the CAR condition (Theorem 3.2(c)) cannot hold for both $U_1$ and $U_2$ unless $\Pr(X_W \in U_1 \cap U_2)$ is either 0 or 1. For suppose that $\Pr(X_O = U_1) > 0$, $\Pr(X_O = U_2) > 0$, and $0 < \Pr(X_W \in U_1 \cap U_2) < 1$. Without loss of generality, there is some $w_1 \in U_1 - U_2$ and $w_2 \in U_1 \cap U_2$ such that $\Pr(X_W = w_1) > 0$ and $\Pr(X_W = w_2) > 0$. Since observations are accurate, we must have $\Pr(X_O = U_1 \mid X_W = w_1) = 1$. If CAR holds for $U_1$, then we must have $\Pr(X_O = U_1 \mid X_W = w_2) = 1$. But then $\Pr(X_O = U_2 \mid X_W = w_2) = 0$. But since $\Pr(X_O = U_2) > 0$, it follows that there is some $w_3 \in U_2$ such that $\Pr(X_W = w_3) > 0$ and $\Pr(X_O = U_2 \mid X_W = w_3) > 0$. This contradicts the CAR condition. ∎

**Example 3.6:** Suppose that $O = \{U_1, U_2, U_3\}$, and all three observations can be made with positive probability. It turns out that in this situation the CAR condition can hold, but only if (a) $\Pr(X_W \in U_1 \cap U_2 \cap U_3) = 1$ (i.e., all of $U_1$, $U_2$, and $U_3$ must hold), (b) $\Pr(X_W \in (U_1 \cap U_2) - U_3) \cup ((U_2 \cap U_3) - U_1) \cup ((U_1 \cap U_3) - U_2)) = 1$ (i.e., exactly two of $U_1$, $U_2$, and $U_3$ must hold), (c) $\Pr(X_W \in (U_1 - (U_2 \cup U_3)) \cup (U_2 - (U_1 \cup U_3)) \cup (U_3 - (U_2 \cup U_1))) = 1$ (i.e., exactly one of $U_1, U_2$, or $U_3$ must hold), or (d) one of $(U_1 - (U_2 \cup U_3)) \cup (U_2 \cap U_3)$, $(U_2 - (U_1 \cup U_3)) \cup (U_1 \cap U_2)$ or $(U_3 - (U_1 \cup U_2)) \cup (U_1 \cap U_2)$ has probability 1 (either exactly one of $U_1$, $U_2$, or $U_3$ holds, or the remaining two both hold).

We first check that CAR can hold in all these cases. It should be clear that CAR can hold in case (a). Moreover, there are no constraints on $\Pr(X_O = U_i \mid X_W = w)$ for $w \in U_1 \cap U_2 \cap U_3$ (except, by the CAR condition, for each fixed $i$, the probability must be the same for all $w \in U_1 \cap U_2 \cap U_3$, and the three probabilities must sum to 1). Case (b) is the most interesting. Let $V_i$ be the set where exactly two of $U_1, U_2$, and $U_3$ hold, and $U_i$ does not hold, for $i = 1, 2, 3$. Suppose that $\Pr(X_W \in V_1 \cup V_2 \cup V_3) = 1$. Note that, since all three observations can be made with positive probability, at least two of $V_1$, $V_2$, and $V_3$ must have positive probability. If only two of them have positive probability, say $V_1$ and $V_2$, then it immediately follows from the CAR condition that there must be some $\alpha$ with $0 < \alpha < 1$ such that $\Pr(X_O = U_3 \mid X_W = w) = \alpha$, for all $w \in V_1 \cup V_2$ such that $\Pr(X_W = w) > 0$. Thus, $\Pr(X_O = U_1 \mid X_W = w) = 1 - \alpha$ for all $w \in V_2$ such that $\Pr(X_W = w) > 0$, and $\Pr(X_O = U_2 \mid X_W = w) = 1 - \alpha$ for all $w \in V_1$ such that $\Pr(X_W = w) > 0$. If all of $V_1, V_2$, and $V_3$ have positive probability, similar arguments show that the probability of each possible observation must be $1/2$. For example, $\Pr(X_O = U_1 \mid X_W = w) = 1/2$ for all $w \in V_2 \cup V_3$ such that $\Pr(X_W = w) > 0$. In case (c), it should also be clear that CAR can hold. Moreover, $\Pr(X_0 = U_i \mid X_W = w)$ is either 0 or 1, depending on whether $w \in U_i$. Finally, for case (d), suppose that $\Pr(X_W \in U_1 \cup (U_2 \cap U_3)) = 1$. CAR holds iff there exists $\alpha$ such that $\Pr(X_O = U_2 \mid X_W = w) = \alpha$ and $\Pr(X_O = U_3 \mid X_W = w) = 1 - \alpha$ forall $w \in U_2 \cap U_3$ such that $\Pr(X_W = w) > 0$. (Of course, $\Pr(X_O = U_1 \mid X_W = w) = 1$ for all $w \in U_1$ such that $\Pr(X_W = w) > 0$.)

Now we show that CAR cannot hold in any other cases. First suppose that $0 < \Pr(X_W \in U_1 \cap U_2 \cap U_3) < 1$. Choose $w \in U_1 \cap U_2 \cap U_3$ such that $\Pr(X_W = w) > 0$, and let $\Pr(X_O = U_i \mid X_W = w) = \alpha_i$, for $i = 1, 2, 3$. Note that $\alpha_1 + \alpha_2 + \alpha_3 = 1$. Suppose $w' \notin U_1 \cap U_2 \cap U_3$ and $\Pr(X_W = w') > 0$. By the CAR condition, $\Pr(X_O = U_i \mid X_W = w')$ is either $\alpha_i$ or 0, depending on whether $w' \in U_i$ or not. Since $\Pr(X_O = U_1 \mid X_W = w') + \Pr(X_O = U_2 \mid X_W = w') + \Pr(X_O = U_3 \mid X_W = w') = 1$, and at least one of these terms is 0, we get the desired contradiction. Similar arguments give a contradiction in all the other cases; we leave details to the reader. ∎

Gill, van der Laan, and Robins [1997] show that for every finite set $W$ of worlds, every set $O$ of observations, and every distribution $\Pr_O$ over $O$, there is a distribution $\Pr^*$ over $\mathcal{R}$ such that the marginal of $\Pr^*$ over $O$ is $\Pr^*$ and CAR holds. The authors summarize this as "CAR is everything". Our examples show that the CAR condition puts quite severe restrictions on the distribution $\Pr^*$ for which CAR holds.

Given that CAR is so difficult to satisfy, the reader may wonder why there is so much study of the CAR condition in the statistics literature. The reason is that some of the special situations in which CAR holds often arise in missing data and survival analysis problems. Here is an example. Suppose that the set of observations can be written as $O = \cup_{i=1}^{k} O_i$, where each $O_i$ is a partition of $W$. Further suppose that observations are generated by the following process. Some $i$ between 1 and $k$ is chosen according to some arbitrary distribution $\Pr_S$; independently, $w \in W$ is chosen according to $\Pr_W$. Then the agent is shown $U \in O_i$ for the unique $U \in O_i$ such that $w \in U$. Intuitively, the partitions $O_i$ represent the observations that can be made with a particular sensor. Thus, $\Pr_S$ determines the probability that a particular sensor is chosen; $\Pr_W$ determines the probability that a particular world is chosen. It is easy to see that this mechanism induces a distribution on $\mathcal{R}$ for which CAR holds.

The important special case with $O = O_1 \cup O_2$, $O_1 = \{W\}$, and $O_2 = \{\{w\} \mid w \in W\}$ corresponds to a simple missing data problem. Intuitively, either complete information is given, or there is no data at all. In this context, CAR is often called *MAR: missing at random*. In more realistic MAR problems, we may observe a vector with some of its components missing. In such cases the CAR condition often still holds. More generally, Gill, van der Laan, and Robins [1997] show that in several problems of survival analysis, observations are generated according to a *randomized monotone coarsening scheme* under which the CAR condition holds.

## 4 BEYOND OBSERVATIONS OF EVENTS

### 4.1 JEFFREY CONDITIONING

In the previous section, we assumed that the information received is of the form "the actual world is in $U$". But information does not always come in such nice packages. Perhaps the simplest generalization of this is to assume that there is a partition $\{U_1, \ldots, U_n\}$ of $W$ and the agent observes $\alpha_1 U_1; \ldots; \alpha_n U_n$, where $\alpha_1 + \cdots + \alpha_n = 1$. This is to be interpreted as an observation that leads the agent to believe $U_j$ with probability $\alpha_j$, for $j = 1, \ldots, n$. According to Jeffrey conditioning,

$$\Pr(V \mid \alpha_1 U_1; \ldots; \alpha_n U_n)$$
$$= \alpha_1 \Pr(V \mid U_1) + \cdots + \alpha_n \Pr(V \mid U_n).$$

Jeffrey conditioning is defined only if $\alpha_i > 0$ implies that $\Pr(U_i) > 0$; if $\alpha_i = 0$ and $\Pr(U_i) = 0$, then $\alpha_i \Pr(V \mid U_i)$ is taken to be 0. Clearly ordinary conditioning is the special case of Jeffrey conditioning where $\alpha_i = 1$ for some $i$ so, as is standard, we deliberately use the same notation for updating using Jeffrey conditioning and ordinary conditioning.

We now want to determine when updating in the naive space using Jeffrey conditioning is appropriate. Thus, we assume that the agent's observations now have the form of $\alpha_1 U_1; \ldots; \alpha_n U_n$ for some partition $\{U_1, \ldots, U_n\}$ of $W$. (Different observations may, in general, use different partitions.) Just as we did for the case that observations are events (Section 3, first paragraph), we once again assume that the agent's observations are accurate. What does that mean in the present context? We simply require that, conditional on making the observation, the probability of $U_i$ really is $\alpha_i$ for $i = 1, \ldots, n$. That is, for $i = 1, \ldots, n$, we have

$$\Pr(X_W \in U_i \mid X_O = \alpha_1 U_1; \ldots; \alpha_n U_n) = \alpha_i. \quad (1)$$

This clearly generalizes the requirement of accuracy given in the case that the observations are events.

Not surprisingly, there is a generalization of the CAR condition that is needed to guarantee that Jeffrey conditioning can be applied to the naive space.

**Theorem 4.1:** *Fix a probability* $\Pr$ *on* $\mathcal{R}$, *a partition* $\{U_1, \ldots, U_n\}$ *of* $W$, *and probabilities* $\alpha_1, \ldots, \alpha_n$ *such that* $\alpha_1 + \cdots + \alpha_n = 1$. *Let* $C$ *be the observation* $\alpha_1 U_1; \ldots, \alpha_n U_n$. *Then the following are equivalent:*

(a) *If* $\Pr(X_O = C) > 0$, *then* $\Pr(X_W = w \mid X_O = C) = \Pr_W(w \mid \alpha_1 U_1; \ldots; \alpha_n U_n)$.

(b) $\Pr(X_O = C \mid X_W = w) = \Pr(X_O = C \mid X_W \in U_i)$ *for all* $i = 1, \ldots, n$ *and* $w \in U_i$ *such that* $\Pr(X_W = w) > 0$.

Part (b) of Theorem 4.1 is analogous to part (c) of Theorem 3.2. There are a number of conditions equivalent to (b) that we could have stated, similar in spirit to the conditions in Theorem 3.2. Note that these are even more stringent conditions than are required for conditioning to be appropriate.

Examples 3.5 and 3.6 already suggest that there are not too many nontrivial scenarios where applying Jeffrey conditioning to the naive space is appropriate. However, just as for the original CAR condition, there do exist special situations in which generalized CAR is a realistic assumption. For ordinary CAR, we mentioned the situation where the set of observations $O$ is a union of partitions, and a specific partition is chosen independently of the process realizing the "actual world" $w$ (see the end of Section 3). For Jeffrey conditioning, a similar mechanism may be a realistic model in some situations where all observations refer to the same partition $\{U_1, \ldots, U_n\}$ of $W$. We now describe a scenario for such a situation. Suppose $O$ consists of $k > 1$ observations $C_1, \ldots, C_k$ with $C_i := \alpha_{i1} U_1; \ldots; \alpha_{in} U_n$ such that all $\alpha_{ij} > 0$. Now, fix $n$ (arbitrary) conditional distributions $\Pr_j$, $j = 1, \ldots, n$, on $W$. Intuitively, $\Pr_j$ is $\Pr_W(\cdot \mid U_j)$. Consider the following mechanism: first an observation $C_i$ is chosen (according to some distribution $\Pr_O$ over $O$); then a set $U_j$ is chosen with probability $\alpha_{ij}$ (i.e., according to the distribution induced by $C_i$); finally, a world $w \in U_j$ is chosen according to $\Pr_j$.

If the observation $C_i$ and world $w$ are generated this way, then the generalized CAR condition holds, that is, conditioning in the sophisticated space coincides with Jeffrey conditioning.

**Proposition 4.2:** *Consider a partition* $\{U_1, \ldots, U_n\}$ *of* $W$ *and a set of* $k$ *observations* $O$ *as above. For every distribution* $\Pr_O$ *over* $O$ *with* $\Pr_O(C_i) > 0$ *for all* $i \in \{1, \ldots, k\}$, *there exists a distribution* $\Pr$ *over* $\mathcal{R}$ *such that* $\Pr_O$ *is the marginal of* $\Pr$ *on* $O$ *and* $\Pr$ *satisfies the generalized CAR condition (part (b) of Theorem 4.1).*

Proposition 4.2 demonstrates that, even though the analogue of the CAR condition expressed in Theorem 4.1 is hard to satisfy in general, at least if the set $\{U_1, \ldots, U_n\}$ is the same for all observations, then there exist *some* priors

$\Pr$ on $\mathcal{R}$ for which the CAR-analogue *is* satisfied for all observations. Below we will see that, in the case of MRE updating, this is not the case any more.

## 4.2   MRE UPDATING

What about cases where the constraints are not in the special form where Jeffrey's conditioning can be applied? Perhaps the most common approach in this case is to use MRE. Given a constraint (where a constraint is simply a set of probability distributions—intuitively, the distributions satisfying the constraint) and a prior distribution $\Pr$, the idea is to pick, among all distributions satisfying the constraints, the one that is "closest" to the prior distribution, where the "closeness" of $\Pr'$ to $\Pr$ is measured using relative entropy. The *relative entropy between* $\Pr'$ *and* $\Pr$ [Kullback and Leibler 1951; Cover and Thomas 1991] is defined as

$$\sum_{w \in W} \Pr'(w) \log \left( \frac{\Pr'(w)}{\Pr(w)} \right).$$

(The logarithm here is taken to the base 2; if $\Pr'(w) = 0$ then $\Pr'(w) \log(\Pr'(w) / \Pr(w))$ is taken to be 0. This is reasonable since $\lim_{x \to 0} x \log(x/c) = 0$ if $c > 0$.) The relative entropy is finite provided that $\Pr'$ is *absolutely continuous* with respect to $\Pr$, in that if $\Pr(w) = 0$, then $\Pr'(w) = 0$, for all $w \in W$. Otherwise, it is defined to be infinite.

The constraints we consider here are all closed and convex sets of probability measures. In this case, it is known that there is a unique distribution that satisfies the constraints and minimizes the relative entropy. Given constraints $C$ and a prior $\Pr$, denote the distribution that minimizes relative entropy with respect to $\Pr$ given $C$ as $\Pr(\cdot \mid C)$.

If the constraints have the form to which Jeffrey's Rule is applicable, that is, if they have the form $\{\Pr' : \Pr'(U_i) = \alpha_i, i = 1, \ldots, n\}$ for some partition $\{U_1, \ldots, U_n\}$, then it is well known that the distribution that minimizes entropy relative to a prior $\Pr$ is $\Pr(\cdot \mid \alpha_1 U_1; \ldots; \alpha_n U_n)$ (see, e.g., [Diaconis and Zabell 1986]). Thus, MRE updating generalizes Jeffrey conditioning (and hence also standard conditioning).

To study MRE updating in our framework, we assume that the observations are now arbitrary closed convex constraints on the probability measure. Again, we assume that the observations are accurate in that, conditional on making the observation, the constraints hold. For now, we focus on the simplest possible case that cannot be handled by Jeffrey updating. In this case, constraints (observations) still have the form $\alpha_1 U_1; \ldots; \alpha_n U_n$, but now the $U_i$'s do not have to form a partition (they may overlap and/or not cover $W$) and the $\alpha_i$ do not have to sum to 1. Such an observation is accurate if it satisfies (1), just as before.

We now want an analogue to Theorems 3.2 and 4.1 show-ing under what conditions applying MRE updating in the naive space leads to the same results as conditioning in the sophisticated space. Seidenfeld [1986] shows that, under very weak conditions, no such analogue is possible if the observations have the form "the conditional probability of $U$ given $V$ is $\alpha$" (as is the case in the Judy Benjamin problem). Here we show that even for observations of the much simpler form $\alpha_1 U_1; \ldots; \alpha_n U_n$, unless we can reduce the problem to Jeffrey conditioning (in which case Theorem 4.1 applies), no such analogue is possible in general: if we cannot reduce the problem to Jeffrey conditioning, then MRE updating essentially *almost never* coincides with sophisticated conditioning.

To demonstrate this, we focus on the simplest possible case. Let $O$ consist of two observations (constraints), $C_i = \alpha_{i1} U_1; \alpha_{i2} U_2$, $i = 1, 2$, where $U_1 - U_2$, $U_1 \cap U_2$, $U_2 - U_1$ and $W - (U_1 \cup U_2)$ are all nonempty. We further assume that $\alpha_{11}, \alpha_{12}, \alpha_{21}, \alpha_{22}$ are all in $(0, 1)$. Note that both $C_1$ and $C_2$ refer to the same events $U_1$ and $U_2$.

We say that observation $C = \alpha_1 U_1; \alpha_2 U_2$ is *Jeffrey-like* iff, when MRE updating on one of the constraints $\alpha_1 U_1$ or $\alpha_2 U_2$, the other constraint holds as well. That is, $C$ is Jeffrey-like (with respect to $\Pr_W$) if either $\Pr_W(U_2 \mid \alpha_1 U_1) = \alpha_2$ or $\Pr_W(U_1 \mid \alpha_2 U_2) = \alpha_1$. Suppose that $\Pr_W(U_2 \mid \alpha_1 U_1) = \alpha_2$; then it is easy to show that $\Pr_W(\cdot \mid \alpha_1 U_1) = \Pr_W(\cdot \mid \alpha_1 U_1; \alpha_2 U_2)$.

Intuitively, if the "closest" distribution $\Pr$ to $\Pr_W$ that satisfies $\Pr(U_1) = \alpha_1$ also satisfies $\Pr(U_2) = \alpha_2$, then $\Pr$ is the closest distribution to $\Pr_W$ that satisfies the constraint $C = \alpha_1 U_1; \alpha_2 U_2$. Note that MRE updating on $\alpha U$ is equivalent to Jeffrey conditioning on $\alpha U; (1-\alpha)(W - U)$. Thus, if $C$ is Jeffrey-like, then updating with $C$ is equivalent to Jeffrey updating. The following theorem shows that, in general, if $C$ is not Jeffrey-like, then there may be no distribution $\Pr$ over $\mathcal{R}$ such that MRE updating coincides with conditioning in the sophisticated space; thus, there can be no equivalent to the CAR condition.

**Theorem 4.3:** *Let* $\Pr$ *be a distribution over* $\mathcal{R}$ *with* $O = \{C_1, C_2\}$ *and* $\Pr(X_O = C_1)$, $\Pr(X_O = C_2) > 0$. *Let* $\Pr^i = \Pr(\cdot \mid \mathcal{R}[(C_i)])$, *and let* $\Pr_W^i$ *be the marginal of* $\Pr^i$ *on* $W$. *If either* $C_1$ *or* $C_2$ *is not Jeffrey-like, then* $\Pr_W^i \neq \Pr_W(\cdot \mid C_i)$, *for* $i = 1, 2$.

We can think of each possible observation $\alpha_1 U_1; \alpha_2 U_2$ (for fixed $U_1$ and $U_2$) as a vector in the set $[0, 1]^2$. Clearly the set of all Jeffrey-like observations is a subset of 0 (Lebesgue) measure of this set. Thus, the set of observations for which MRE conditioning corresponds to conditioning in the sophisticated space is a (Lebesgue) measure 0 set in the space of possible observations.

Theorem 4.3 shows that, in the case where only two observations are possible, MRE cannot coincide with conditioning in the sophisticated space unless both observa-tions are Jeffrey-like. If we allow an arbitrary number of observations rather than just two, then there may be some very special non-Jeffrey-like combinations of priors $\Pr$ and observations such that MRE updating corresponds to conditioning in the sophisticated space. However, in marked contrast to the case for Jeffrey conditioning, these remain isolated cases. More specifically, Proposition 4.2 shows that, given an arbitrary set $O$ of observations to which Jeffrey conditioning can apply, where all the observations in $O$ refer to the same events, and a distribution $\Pr_O$ on $O$, we can *always* construct *some* distribution $\Pr$ over $\mathcal{R}$ such that $\Pr(X_O = C) = \Pr_O(C)$ for all $C \in O$ and $\Pr$ satisfies the generalized CAR condition. Proposition 4.4 shows that, if the $U_i$ are allowed to overlap, this is not possible in general. We first need some terminology. For given sets $W$ and $O = \{C_1, \ldots, C_k\}$, we say $(\lambda_1, \ldots, \lambda_k)$ is *CAR-compatible* iff there exists a distribution $\Pr$ over $\mathcal{R}$ with $\Pr(X_O = C_i) = \lambda_i$ such that the generalized CAR condition holds. We let $\Delta^k$ stand for the unit simplex in $\mathbf{R}^k$.

**Proposition 4.4:** *Fix an (arbitrary) set* $\{U_1, \ldots, U_n\}$ *of subsets of* $W$ *such that* $(U_1 \cap U_2) - \cup_{i=3..n} U_i$, $(U_1 - U_2) - \cup_{i=3..n} U_i$ *and* $(U_2 - U_1) - \cup_{i=3..n} U_i$ *are all nonempty. Suppose that there are* $k > 1$ *possible observations,* $C_1, \ldots, C_k$, *with* $C_i := \alpha_{i1} U_1; \ldots; \alpha_{in} U_n$, *such that all* $\alpha_{ij} > 0$. *Define*

$$S := \{(\lambda_1, \ldots, \lambda_k) \in \Delta^k : \lambda \text{ is CAR-compatible.}\}.$$

*Then* $S$ *is a subset of* $\Delta^k$ *of Lebesgue measure* 0.

Proposition 4.4 says that for all sets of observations $O$ of a certain kind, *almost all* (that is, a measure 1 subset of) priors over observations are such that CAR cannot hold. To compare this with Theorem 4.3, note first that whether a constraint $C_i = \alpha_{i1} U_1; \alpha_{i2} U_2$ is Jeffrey-like depends not only on the $\alpha_{ij}$ but also on the marginal prior distribution $\Pr_W$ over $W$. Theorem 4.3 says that for all sets of observations of a certain kind, *all* priors over worlds are such that for *almost all observations*, CAR cannot hold.

## 5 DISCUSSION

We have studied the circumstances under which conditional updating, Jeffrey conditioning, and MRE updating in a naive space can be justified, where "justified" for us means "agrees with conditioning in the sophisticated space". The main message of this paper is that, except for quite special cases, the three methods cannot be justified. Figure 1 summarizes the main insights of this paper in more detail.

As we mentioned in the Introduction, the idea of comparing an update rule in a "naive space" with conditioning in a "sophisticated space" is not new; it appears in the CAR literature and the MRE literature (as well as in papers such as [Halpern and Tuttle 1993] and [Dawid and Dickey 1977]).

| observation type | set of observations $O$ | simplest applicable update rule | when it coincides with sophisticated conditioning |
|---|---|---|---|
| event | pairwise disjoint | naive conditioning | iff CAR holds (Theorem 3.2) |
| event | arbitrary set of events | naive conditioning | iff CAR holds (Theorem 3.2) |
| probability vector | probabilities of partition | Jeffrey conditioning | iff generalization of CAR holds (Theorem 4.1) |
| probability vector | probabilities of overlapping sets | MRE | no general characterization |

Figure 1: Conditions under which updating in the naive space coincides with conditioning in the sophisticated space.

In addition to bringing these two strands of research together, our own contributions are the following: (a) we show that the CAR framework can be used as a general tool to demistify paradoxes of conditional probability; (b) we show that the CAR condition has a natural extension to cases where Jeffrey conditioning can be applied (Theorem 4.1); (c) we show that no CAR-like condition can exist in general for cases where only MRE (and not Jeffrey) updating can be applied (Theorem 4.3).

Our results suggest that working in the naive space is rather problematic. On the other hand, as we observed in the introduction, working in the sophisticated space (even assuming it can be constructed) is problematic too. So what are the alternatives?

For one thing, it is worth observing that MRE updating is not always so bad. In many successful practical applications, the "constraint" on which to update is of the form $\frac{1}{n}\sum_{i=1}^{n} X_i = t$ for some large $n$, where $X_i$ is the $i$th outcome of a random variable $X$ on $W$. That is, we observe an empirical average of outcomes of $X$. In such a case, the MRE distribution is "close" (in the appropriate distance measure) to the distribution we arrive at by sophisticated conditioning. That is, if $\Pr'' = \Pr_W(\cdot \mid E(X) = t)$, $\Pr' = \Pr(\cdot \mid \mathcal{R}[\langle \frac{1}{n}\sum_{i=1}^{n} X_i = t\rangle])$, and $Q^n$ denotes the $n$-fold product of a probability distribution $Q$, then for sufficiently large $n$, we have that $(\Pr'')^n \approx (\Pr'_W)^n$ [van Campenhout and Cover 1981; Grünwald 2001]. Thus, in such cases MRE (almost) coincides with sophisticated conditioning after all. (See [Dawid 2001] for a discussion of how this result can be reconciled with the results of Section 4.)

But when this special situation does not apply, it is worth asking whether there exists an approach for updating in the naive space that can be easily applied in practical situations, yet leads to *better*, in some formally provable sense, updated distributions than the methods we have considered? A very interesting candidate, often informally applied by human agents, is to simply *ignore* the available extra information. It turns out that in many situations this update rule behaves better, in a precise sense, than the three methods we have considered. This will be explored in future work.

Our discussion here has focused completely on the probabilistic case. However, these questions also make sense for other representations of uncertainty. Interestingly, in [Friedman and Halpern 1999], it is shown that AGM-style belief revision [Alchourrón, Gärdenfors, and Makinson 1985] can be represented in terms of conditioning using a qualitative representation of uncertainty called a *plausibility measure*; to do this, the plausibility measure must satisfy the analogue of Theorem 3.2(a), so that observations carry no more information than the fact that they are true. No CAR-like condition is given to guarantee that this condition holds for plausibility measures though. It would be interesting to know if there are analogues to CAR for other representations of uncertainty, such as *possibility measures* [Dubois and Prade 1990] or *belief functions* [Shafer 1976].

## Acknowledgments

We thank the referees of the UAI submission for their perceptive comments. The first author was supported by a travel grant awarded by the Netherlands Organization for Scientific Research (NWO). The second author was supported in part by NSF under grant IIS-0090145, by ONR under grants N00014-00-1-0341, N00014-01-1-0511, and N00014-02-1-0455, by the DoD Multidisciplinary University Research Initiative (MURI) program administered by the ONR under grants N00014-97-0505 and N00014-01-1-0795, and by a Guggenheim and a Fulbright Fellowship. Sabbatical support from CWI and the Hebrew University of Jerusalem is also gratefully acknowledged.

## References

Alchourrón, C. E., P. Gärdenfors, and D. Makinson (1985). On the logic of theory change: partial meet functions for contraction and revision. *Journal of Symbolic Logic 50*, 510–530.

Bar-Hillel, M. and R. Falk (1982). Some teasers concerning conditional probabilities. *Cognition 11*, 109–122.

Cover, T. M. and J. A. Thomas (1991). *Elements of Information Theory*. New York: Wiley.

Csiszár, I. (1991). Why least squares and maximum entropy? An axiomatic approach to inference for linear inverse problems. *Annals of Statistics 19*(4), 2032–2066.

Dawid, A. (2001, January). A note on maximum entropy and Bayesian conditioning. Unpublished manuscript.

Dawid, A. and J. Dickey (1977). Likelihood and Bayesian inference from selectively reported data. *Journal of the American Statistical Association 72*(360), 845–850.

Diaconis, P. and S. L. Zabell (1986). Some alternatives to Bayes's rule. In B. Grofman and G. Owen (Eds.), *Proc. Second University of California, Irvine, Conference on Political Economy*, pp. 25–38.

Dubois, D. and H. Prade (1990). An introduction to possibilistic and fuzzy logics. In G. Shafer and J. Pearl (Eds.), *Readings in Uncertain Reasoning*, pp. 742–761. San Francisco: Morgan Kaufmann.

Friedman, K. and A. Shimony (1971). Jaynes' maximum entropy prescription and probability theory. *Journal of Statistical Physics 9*, 265–269.

Friedman, N. and J. Y. Halpern (1997). Modeling belief in dynamic systems. Part I: Foundations. *Artificial Intelligence 95*(2), 257–316.

Friedman, N. and J. Y. Halpern (1999). Modeling belief in dynamic systems. Part II: Revision and update. *Journal of A.I. Research 10*, 117–167.

Gardner, M. (1961). *Second Scientific American Book of Mathematical Puzzles and Diversions*. Simon & Schuster.

Gill, R., M. van der Laan, and J. Robins (1997). Coarsening at random: Characterisations, conjectures and counter-examples. In *Proceedings First Seattle Conference on Biostatistics*, pp. 255–294.

Grove, A. J. and J. Y. Halpern (1997). Probability update: conditioning vs. cross-entropy. In *Proc. Thirteenth Conference on Uncertainty in Artificial Intelligence (UAI '97)*, pp. 208–214.

Grünwald, P. (2001). Strong entropy concentration, game theory and algorithmic randomness. In *Proceedings of the Fourteenth Annual Conference on Computational Learning Theory*, pp. 320–336.

Halpern, J. Y. and R. Fagin (1989). Modelling knowledge and action in distributed systems. *Distributed Computing 3*(4), 159–179.

Halpern, J. Y. and M. R. Tuttle (1993). Knowledge, probability, and adversaries. *Journal of the ACM 40*(4), 917–962.

Heitjan, D. and D. Rubin (1991). Ignorability and coarse data. *Annals of Statistics 19*, 2244–2253.

Jeffrey, R. C. (1968). Probable knowledge. In I. Lakatos (Ed.), *International Colloquium in the Philosophy of Science: The Problem of Inductive Logic*, pp. 157–185. North Holland Publishing Co.

Kleinbaum, D. (1999). *Survival Analysis: A self-Learning Text*. Statistics in the Health Sciences. Springer-Verlag.

Kullback, S. (1959). *Information Theory and Statistics*. Wiley.

Kullback, S. and R. A. Leibler (1951). On information and sufficiency. *Annals of Mathematical Statistics 22*, 76–86.

Mosteller, F. (1965). *Fifty Challenging Problems in Probability with Solutions*. Reading, Mass.: Addison-Wesley.

Nielsen, S. (1998). *Coarsening at Random and Simulated EM Algorithms*. Ph. D. thesis, Department of Theoretical Statistics, University of Copenhagen.

Rubin, D. (1976). Inference and missing data. *Biometrika 63*, 581–592.

Seidenfeld, T. (1986). Entropy and uncertainty. *Philosophy of Science 53*, 467–491.

Shafer, G. (1976). *A Mathematical Theory of Evidence*. Princeton, N.J.: Princeton University Press.

Shafer, G. (1985). Conditional probability. *International Statistical Review 53*(3), 261–277.

Shore, J. E. and R. W. Johnson (1980). Axiomatic derivation of the principle of maximum entropy and the principle of minimimum cross-entropy. *IEEE Transactions on Information Theory IT-26*(1), 26–37.

van Campenhout, J. and T. Cover (1981). Maximum entropy and conditional probability. *IEEE Transactions on Information Theory IT-27*(4), 483–489.

van Fraassen, B. C. (1981). A problem for relative information minimizers. *British Journal for the Philosophy of Science 32*, 375–379.

vos Savant, M. (1990). Ask Marilyn. *Parade Magazine*, 15. There were also followup articles in *Parade Magazine* on Dec. 2, 1990 (p. 25) and Feb. 17, 1991 (p. 12).